\documentclass{article}
\usepackage{neurips_2022}

\usepackage[utf8]{inputenc} 
\usepackage[T1]{fontenc}    
\usepackage{hyperref}       
\usepackage{url}            
\usepackage{booktabs}       
\usepackage{amsfonts}       
\usepackage{nicefrac}       
\usepackage{microtype}      
\usepackage{xcolor}         
\usepackage{comment}
\usepackage{algorithm}
\usepackage[backend=biber,style=ieee]{biblatex}
\usepackage{algpseudocode}
\usepackage{amssymb} 
\title{Fair Differentially Private Federated \\ Learning Framework}

\begin{document}

\maketitle

\begin{abstract}

Federated learning (FL) is a distributed machine learning strategy that enables participants to collaborate and train a shared model without sharing their individual datasets. Privacy and fairness are crucial considerations in FL. While FL promotes privacy by minimizing the amount of user data stored on central servers, it still poses privacy risks that need to be addressed. Industry standards such as differential privacy, secure multi-party computation, homomorphic encryption, and secure aggregation protocols are followed to ensure privacy in FL. Fairness is also a critical issue in FL, as models can inherit biases present in local datasets, leading to unfair predictions. Balancing privacy and fairness in FL is a challenge, as privacy requires protecting user data while fairness requires representative training data. This paper presents a "Fair Differentially Private Federated Learning Framework" that addresses the challenges of generating a fair global model without validation data and creating a globally private differential model. The framework employs clipping techniques for biased model updates and Gaussian mechanisms for differential privacy. The paper also reviews related works on privacy and fairness in FL, highlighting recent advancements and approaches to mitigate bias and ensure privacy. Achieving privacy and fairness in FL requires careful consideration of specific contexts and requirements, taking into account the latest developments in industry standards and techniques.
 
\end{abstract}

\section{Introduction}

Federated learning (FL) is a distributed machine learning strategy that enables numerous participants to work together and train a standard model without disclosing their individual data sets. Due to its ability to circumvent the drawbacks of centralized learning, including data privacy problems, network bandwidth restrictions, and data heterogeneity, this approach has attracted much attention in recent years.  

The objective of FL is to keep the training dataset in the location where it was created and to utilize the on-device processing power to train the models at each learning device in the federation. Each learning device transfers its local model parameters, rather than the raw training dataset, to an aggregating unit after training a local model. The global model is updated by the aggregator using the local model parameters, and the updated global model is then supplied back to the individual local learners for their usage. As a result, without explicitly accessing their privacy-sensitive data, each local learning device benefits from the datasets of the other learners exclusively through the global model given by the aggregator. But his idea can cause a breach in data privacy because sometimes sensitive data can be passed through a training loop.  

FL promotes privacy by default by reducing the amount of user data that is stored on the network's central server. In other words, rather than sharing their actual data, individuals are asked to submit local training model parameters. The adversary can partially reveal each participant's training data in the original training dataset based on their model updates. Therefore, FL still poses some privacy issues, according to recent research. These serious dangers in FL can be categorized into various inference-based attack types.

Hence privacy is a key concern in federated learning, and several industry standards are followed to ensure the participants' privacy. One of the most widely accepted industry standards for privacy in federated learning is differential privacy. Differential Privacy (DP) is a mathematical framework that provides a rigorous definition of privacy. It has been widely adopted in the research community to ensure participants' privacy in federated learning. Another industry standard for privacy in federated learning is secure multi-party computation (MPC) techniques. Secure MPC techniques enable multiple parties to perform computations on their private data while keeping their data confidential. This ensures that sensitive information is not revealed to other parties in the collaboration. Other industry standards for privacy in federated learning include encryption techniques such as homomorphic encryption and secure aggregation protocols, which enable parties to securely aggregate their local model updates without revealing their private data. Overall, the industry standards for privacy in federated learning constantly evolve as new research is conducted. It is essential to stay up-to-date with the latest developments to ensure participants' privacy in federated learning collaborations.

Fairness is a critical issue in machine learning and equally crucial in federated learning. Federated learning (FL) models can inherit the biases present in the local datasets of each participating party, leading to unfair predictions and decisions. In other words, if there are n number of devices and some of them contain or generate biased data, then this will affect the global model. Thus, ensuring fairness in federated learning is crucial to prevent discriminatory outcomes in the global model and promote equal opportunities for all individuals.

\textbf{Privacy vs  Fairness}: In the context of federated learning, it is a challenge to balance fairness and privacy. On the one hand, the model must perform equally well for all users, independent of their demographic details like race, gender, age, or location. Contrarily, privacy mandates that users' data not be shared or utilized in a manner that jeopardizes their privacy. To achieve fairness in federated learning, it is necessary to ensure that the training data used by each node is representative of the overall population. This can be accomplished by including data from various sources or employing strategies like stratified sampling, in which the data is divided into groups depending on particular characteristics. At the same time, to preserve privacy, federated learning uses several techniques, such as encryption, differential privacy, and secure multi-party computation, to ensure that the raw data is not shared with the central server or other nodes. This is critical for ensuring user trust and compliance with data protection laws. In conclusion, balancing fairness and privacy in federated learning requires a careful approach that considers each use case's specific context and requirements.


\textbf{Main contribution}: This paper highlights the problem of generating a fair global model without validation data.  We also address the creation of a globally private differential model. We have put forth a "Fair Differentially Private Federated Learning Framework" to address these issues. Our framework employs clipping techniques to handle biased model updates and Gaussian mechanics for differential privacy.

\section{Related Works}
\label{gen_inst}
This section discusses about the recent works being done related to privacy and fairness in Federated learning. 
Kairouz et al.~\cite{kairouz2021advances} discusses the recent advancements in Federated Learning by providing an extensive collection of open problems and challenges in the federated framework. it describes the open challenges related to relaxing the core assumptions of FL, improving its efficiency, providing some privacy guarantees with defensive mechanisms against attacks, and ensuring fairness in federated settings. Our paper discusses about the challenges of providing privacy of user data in a federated setting along with mitigating bias which may occur due to data heterogeneity. 

Achieving fairness in Federated Learning (FL) is challenging because mitigating bias inherently requires using the sensitive attribute values of all clients. At the same time, FL aims to protect privacy by not giving access to the client's data. Pentyala et al.~\cite{pentyala2022privfairfl} have solved this conflict between privacy and fairness in FL by combining FL with Differential Privacy (DP) and Secure Multi-Party Computation (MPC). To do this, the authors propose an approach for fair machine learning models under privacy guarantees without asking the clients to disclose their sensitive attribute values. Zhou et al. mitigates the bias by requiring access to the data of all clients. They proposed pre-processing and in-processing methods to mitigate bias even with heterogeneous data. They evaluated their proposed approach by analyzing the effects of different data distributions on model's performance as well as on fairness metrics. Abay et al.~\cite{abay2020mitigating} studied the cause of bias in federated learning and proposes three methods to mitigate them. They demonstrate that by utilizing local reweighing produces fair models without sacrificing privacy or model accuracy, even when only a small fraction of parties engage in the fairness procedure.

In machine learning models, information about data distribution can leverage the performance of a model. Du et al.~\cite{du2021fairness} have shown that statistics about data distribution across clients in the FL framework can help in improving the utility of a model, or can make them for fair in the case of imbalanced and non-i.i.d. data. But this will lead to privacy leakage in FL settings. One common solution nowadays to protect the input data in FL settings is to aggregate data with MPC ~\cite{ezzeldin2021fairfed} ~\cite{zhang2020fairfl}. However, there are still chances of information leakage using MPC-based aggregation on the client's side. So, to avoid this leakage, DP-based methods work better that ensures privacy guarantees by adding noise to the aggregated values. 

    One of the primary motivation behind the Federated Learning framework is to provide privacy guarantees. Differential privacy has been considered as an industrial standard for the notion of privacy. There are some recent works that ensure client-side differential privacy by clipping client's transmitted model updates before adding any noise. Xinwei et al.~\cite{zhang2022understanding} worked on the similar principle by proposing a clipped FedAvg model which resulted well even with significant heterogeneous data. They also provided a convergence analysis depicting the relationship between clipping bias and the distribution of clients updates. Geyer et al. ~\cite{geyer2017differentially} aimed to hide the clients' contribution while training the federated learning model by managing the trade-off between utility and privacy. To do so, the authors approximated the clients' model updates with a randomized mechanism so that a single client's contribution can be hidden in the aggregation. They experimented using MNIST data sets in order to test the proposed approach and showed that client level differential privacy is feasible with high model accuracies. 

On one hand federated learning is suitable for privacy-protection, but on the other hand this makes FL more susceptible to adversarial attacks due to its decentralized settings. A possible attack in federated settings is backdoor attack, in which to desire a misclassification, an adversary embeds a backdoor functionality to a model while training and can be activated later. Ozdayi et al. ~\cite{ozdayi2021defending} proposed a defensive mechanism against backdoor attack in Federated Learning in which they try to adjust the aggregation server's learning rate, per dimension and per round based on the model updates. They clipped the corrupted model updates before sending the model updates to the aggregated server.  This technique can be further extended achieving fairness in federated learning by clipping the biased model updates which is described below in the proposed work.

\section{Background Study}
\label{headings}
\subsection{Federated Learning}
\par In the year 2016, a novel concept of federated learning(FL) has been proposed by Google \cite{mcMahan} allowing a collaborative model to be built from distributed data without sharing it thereby preserving privacy and security of the data. Assuming $M$ data owners $\{D_1,~D_2,~\dots,~D_M\}$, Yang \textit{et al.} \cite{flbook} define federated learning as a process of constructing collaborative model $M_{Fed}$ with accuracy $A_{Fed}$ such that
\begin{equation}\label{eq:defFL}
|A_{Fed} - A_{Cen}| < \delta
\end{equation}
where $A_{Cen}$ is the accuracy of centralized machine learning on
data $D=D_1\cup D_2 \cup \dots \cup D_M$ and $\delta$ is a non-negative real number. The federated learning algorithm is said to have $\delta$-accuracy loss \cite{flbook}. The terms node and client are used interchangeably to represent data owner. Mostly, two categories of federated learning are being investigated - one, where data at clients have the same features but different samples, called horizontal federated learning, and second, called vertical federated learning, where clients have different feature spaces. In our proposed solution, we intend to work with horizontal federated learning. 
\par Federated Learning considers the following optimization problem\cite{zhang2022understanding}:
 \begin{equation}
     min(x)[f(x)\triangleq\sum_{i}^{N}f_{i}(x)]=\mathit{\mathbb{E}}_{\xi{}\sim D_{i}}F(x;\xi)
  \end{equation}
where N is the total number of clients. ith client optimizes on a local model $f_{i}$ that is the expectation of loss function $F(x;\xi)$. $x_{i}$ denotes the model parameters and $\xi$ is the data sample, and the expectation is taken over the local dataset $D_{i}$.
\par In every communication round, the server transmits the global model parameters to the selected clients. These clients perform the local model training on their own individual dataset and send their updated parameters to the server which then aggregates the differences to the global model. This communication stops when convergence is achieved.
\subsection{Differential Privacy}
In our work, we intend to propose a framework that is subject to the privacy guarantees of Differential Privacy(DP)\cite{dwork}. The formal definition of differential privacy is given as:
\par An algorithm M is said to be $\left( \varepsilon,\delta \right)$ - differentially private if
    \begin{equation}
    P(M(D\in S)\le e^\in P(M(D')\in S)+\delta
    \end{equation}
   where $D$ and $D'$ are neighbouring datasets and $S$ is an arbitrary subset of outputs of $M$. A smaller value of $\varepsilon$ enforces stronger privacy.
\par Considering the FL setup, there are two main definitions of DP are used in algorithm design:
\begin{itemize}
\item Sample-level Differential Privacy (SL-DP)- SL-DP is similar to the centralized DP which protects only one local sample so that it is not possible for the central server to identify one sample from the union of all the local data distributions. 
\item Client-level Differential Privacy(CL-DP)- CL-DP is a more strict notion of DP in FL.  In CL-DP, the central server must not be able to identify the participation of a single client by looking at the local model updates.
\end{itemize}
\par Applying differential privacy in the federated setting is a complex task when compared to the centralized setting because of different data distributions among the clients. Chen et al\cite{chen2020} in their recent work have shown how the different data distributions even in the centralized setting affect the performance of DP-SGD. So, the impact of the heterogeneous data distribution in a federated learning algorithm that protects DP is quite unclear. Though there are many alternative ways in which we can apply DP to federated learning but none of them have a rigorous theoretical guarantee. 
\subsection{Clipping}
It is quite well-known that exploding gradients are a common problem while training deep neural networks. To solve this issue gradient clipping is an effective solution\cite{goodfellow}. In the gradient clipping method, a maximum threshold value is set for the gradients in the training process. If they exceed that threshold value, they will be clipped down to the threshold. This ensures that these gradients are in a defined range and do not update too drastically in the training. Clipping can be generically implemented in two ways:
\begin{itemize}
    \item Gradient Clipping by value- It is the simplest way of clipping the gradients. We set a minimum and maximum limit for the gradients in clipping by the value method. If the gradient exceeds the upper limit or is less than the lower limit then it is clipped to the threshold values.
    \item Gradient Clipping by norm- It is also similar to the clipping by value method. The only difference is that in clipping by norm we clip the gradients by multiplying the unit vector of the gradients with the threshold.
\end{itemize}
\par Clipping is the main operation in providing DP guarantees in FL algorithms. Hence, it is quite important to understand how clipping can affect the convergence of an FL algorithm.
\subsection{Fairness}
Machine learning algorithms, in recent times, are being extensively used for automated decision-making tasks as they have the benefit of processing a large amount of data, and their outputs seem to be fair but unfortunately, this is not the case. Algorithms are vulnerable to biases that render their decisions “unfair”\cite{verma2019}. A biased model may inadvertently encode human prejudice due to biases in data\cite{mehrabi2021}. An algorithm is said to be fair if it does not discriminate against groups of individuals based on any sensitive attributes, such as race and gender inherently or in the decision-making process. 
\par In the absence of validation data, it has been found that federated learning can introduce new sources of bias through various data generation processes and assumptions. One such fairness challenge in federated learning settings is collaborative fairness\cite{ezzeldin2021fairfed}. It is defined as rewarding a client with a high contribution and a better performing local model more than a low contribution client. Hence, there is a need to propose a fair federated framework that can overcome these challenges.
\section{Proposed Solutions}

In federated learning environment, for each round of communication between central server and clients, clients send their model updates to a central server. Central server chooses some client updates randomly for aggregation i.e., let $K$ be the number of clients and central server randomly selects $m_t \leq K$ clients at time step $t$ then the global weight $w_{t+1}$ is computed by aggregating model updates from these $m_t$ clients. Due to the aggregation, the central server may incorporate the biased updates from client devices to global model. This may result in the biased inference for all the clients due to the biased update from some. Also, we must protect each clients data i.e, we must ensure that the global model does not reveal whether a client participated or not during the training procedure or not. To overcome these problems, in this paper we propose 'Fair Differentially Private Federated Learning Framework'. In our framework, we clip the  biased model updates and add a randomized Gaussian noise in order to hide a single client's contribution in the aggregation. First we focus on the choice of the clipping approaches, two major clipping approaches used in the literature are as follows.

\begin{enumerate}
    \item \textbf{Model clipping:} In this approach clients directly clip the model updates sent to the server i.e., if $clip()$ is a clipping function with threshold $c$, then in model clipping we use $clip(x_i^{t}, c)$. This approach suffers from large clipping threshold which might require large perturbation for differential privacy.
    \item  \textbf{Difference clipping:} In this approach, the clients clips the difference between the initial model model and the model after training on their local data i.e. $\triangle w_t \leftarrow w_t - w_{t-1}$. Note that to do so, the clients must store the initial model, compute the difference between the learnt and initial model and communicate it to the server. Relatively, the clipping threshold is smaller than the model clipping approach and hence less perturbation is required to satisfy DP. 
\end{enumerate}

Our clipping operator is inspired from the differentially private clipping for model update in \cite{geyer2017differentially}. The clipping operation in \cite{geyer2017differentially} focuses on clipping for DP. We introduce another term to avoid any biased updates. Our clipping operator is as follows.

\begin{equation} \label{proposed}
    w_{t+1} = w_t + \frac{1}{m_t} \sum_{i=0}^{m_t-1} \frac{\triangle w_t}{max(1, \frac{||\triangle w_i||_2}{S}, \frac{||\triangle w_i||_2}{M})} + N (0, \sigma^2S^2)
\end{equation}

Here, $w_{t+1}$ represents the aggregated weight from randomly chosen $m_t$ clients. The $1/m_t$ averages the clipped updates from each client and $||\triangle w_i||_2/S$ represents the clipping for differentially private aggregation while $||\triangle w_i||_2/M$ represents the clipping to avoid biased paramter updates using some predefined threshold $M$. The Gausian noise $N (0, \sigma^2S^2)$ avoids the membership inference for each client. Algorithm \ref{prop_algo} present our proposed framework for updating global differentially private clipped model in each communication round. 

\begin{algorithm} \label{prop_algo}
    \caption{Fair Differentially Private Federated Learning Framework}\label{fair-dp-fed}
    \begin{algorithmic}
        \State Initialize: $w_i^0=w^0, i=1,2,...,K$
        \For{$t=0,1,...,T-1$ (number of communication rounds)}
            \For{$i=0,1,...,m_t \in K$ compute gradient for all clients in parallel}
                \State Update model for $w_i^{t,0}=w_t$
                \For{$j=0,1...epochs$ (train the received model on local data)}
                    \State Compute gradient $\nabla_g$ on local data
                    \State Perform $w_i^{t,j+1}=w_i^{t,j} - \eta \nabla_g$
                \EndFor
                \State Compute difference $\triangle w_i^t = w_i^{t,epochs}-w_i^{t,0}$
                \State Communicate $\triangle w_i^t$ to global server
            \EndFor
            \State Perform global averaging $w^{t+1}$ using Eq. \ref{proposed}
        \EndFor
    \end{algorithmic}
\end{algorithm}

\textbf{Choice for $S$}: There exists a trade-off between the noise to be added and minimize the information loss. While we want to protect the membership disclosure for each client but we would also want to maximize the learning. In this case we follow the recommendation given in \cite{abadi2016deep}, we chose the median of the unclipped $\triangle w_i^t$ as the clipping bound for $S$ i.e. $S = median(\triangle w_i^t)_{i \in m_t}$. In comparison with mean, median causes less information loss. It is important here to remind that the first assumption i.e. the number of unbiased clients are significantly more than the biased clients is necessary here.

\textbf{Choice for $M$}: The choice of $M$ in our case is very important and must be decided by an expert in the field. As the difference in the biased updates and unbiased updates can vary between applications. Our simple idea is to ignore all the updates which has higher difference than some threshold $M$, in some complex applications the choice of $M$ can be random as well. 

\textbf{Limitation:}

Some of the limitations of the idea are as follows.

\begin{enumerate}
    \item The choice for the parameter $M$ depends on the human expert. In case of biased human expert the threshold may fail to clip the parameters.
    \item The paper does not present any empirical evidence.
    \item The assumption of having significantly more unbiased clients than biased clients is a strong assumptions. It may not be easy to satisfy in real-world applications. 
\end{enumerate}

\section{Conclusion and Future Work}

In this paper, we highlight the problem of generating fair global model in the absence of validation data. We also deal with the generation of differentially private global model. To overcome these problems, we have proposed a framework called, 'Fair Differentially Private Federated Learning Framework'. Our framework uses different clipping operations to deal with biased model updates and Gaussian mechanishm. We also add gaussian noise to avoid membership inference for client's dataset. 

Even though the study presents an interesting intuitive idea. This paper lacks empirical evidence. We must provide empirical evidence in future studies to support the claim of this paper. Another interesting direction is to see up to what ratio of biased and unbiased clients our framework can provide good results. 

\section{Author's Contributions}

\textbf{Ayush K. Varshney}: Conceptualization, Methodology, Formal analysis, Validation, Visualization, Writing (Proposed Work, Conclusion).

\textbf{Sonakshi Garg}: Methodology, Formal analysis, Validation, Writing (Background Study).

\textbf{Arka Ghosh}: Methodology, Formal analysis, Validation, Writing (Related Works).

\textbf{Sargam Gupta}: Methodology, Formal analysis, Validation, Writing (Introduction).

\printbibliography

\end{document}